\documentclass[runningheads]{llncs}
\usepackage{graphicx}
\usepackage{mathtools}
\begin{document}
\title{Effect of shapes of activation functions on predictability in the echo state network \thanks{This work is partially supported by Toyota Motor Corp. and KAKENHI No. 17H03280.}}
%
%
\author{Hanten Chang\inst{1} \and
Shinji Nakaoka\inst{2} \and
Hiroyasu Ando\inst{1}}
\authorrunning{H. Chang et al.}
%
\institute{Faculty of Engineering, Information and Systems, University of Tsukuba, Japan
\email{s1820554@s.tsukuba.ac.jp,ando@sk.tsukuba.ac.jp}\\
\and
Graduate school of life science, Hokkaido University\\
\email{snakaoka@sci.hokudai.ac.jp}}
\maketitle              
\begin{abstract}
We investigate prediction accuracy for time series of Echo state networks with respect to several kinds of activation functions. As a result, we found that some kinds of activation functions with an appropriate nonlinearity show high performance compared to the conventional sigmoid function.

\end{abstract}
\section{Introduction}\vspace{-2mm}
Machine learning with artificial neural networks is nowadays widely applied to many research fields ranging from engineering to medical and biological sciences. 
Specifically, learning with multi-layered or deep neural networks with feed forward and recurrent structures has gathered attention due to its extremely high performance for several kinds of tasks. 
However, deep learning requires expensive computational cost. 
To  reduce the cost, learning method for recurrent neural network has been focused and developed during the last decade \cite{jaeger2004harnessing,lukovsevivcius2009reservoir}. 
One of the seminal method should be the echo state network (ESN), where connection weights in the network are fixed and only read-out connection weights are varied.  Therefore, the computational cost of ESN is cheap.  Due to its simplicity, ESN is widely applicable to several kinds of tasks such as prediction of chaotic time series 
\cite{jaeger2004harnessing}, image recognition \cite{woodward2011reservoir}, and so on.

As is the case with most artificial neural networks, the sigmoid function is used as the activation function for each unit in the normal ESN, while other functions are also used in order to improve its predictability \cite{wang2006harnessing,woodward2011reservoir}.
In this paper, we focus on the effect of changing shape of the activation function of ESN on its prediction of time series. Specifically, non-monotonic functions such as the sinc function and other even functions are applied to ESN. We numerically investigate prediction accuracy of ESN with those activation functions for several kinds of time series. 

\section{Method}\vspace{-2mm}
\subsection{Echo State Network}\vspace{-1mm}
The dynamics of ESN in this study is described as follows \cite{lukovsevivcius2009reservoir}:  
\begin{align}
X(t) &= f(W^\text{in} [1;U(t)] + WX(t-1)), \label{rule}
\end{align}
where $X(t)\in \mathrm{R}^{D_x}$ is the state of units at time $t$. The number of 
units of ESN is $D_x$.  $U(t)\in \mathrm{R}^{D_u}$ is an input signal to ESN.  Due to the dimension of the input and the state of units,  the weight matrices are represented by $W^\text{in} \in \mathrm{R}^{D_x \times (1+ D_u)}$ and $W \in \mathrm{R}^{D_x \times D_x}$. Finally, $f$ is an activation function applied element-wise.  $[\cdot;\cdot]$ is a vector concatenation. 

The output from ESN is represented by $y(t) \in \mathrm{R}^{D_y}$ and 
defined as 
$
	y(t) = W^\text{out}[1; U(t); X(t)],  
$
where  $W^\text{out} \in \mathrm{R}^{D_y \times (1+ D_u+D_x)}$ is the output weight matrix. In this study, $W^\text{out}$ is estimated  by the ridge regression with the regularization parameter $\lambda = 10^{-8}$.  $W^\text{in}, W$ are randomly determined by following the normal distribution $N(0,\sigma^2)$. 

\vspace{-2mm}
\subsection{Activation functions}
\begin{figure}[t]
\includegraphics[width=\textwidth]{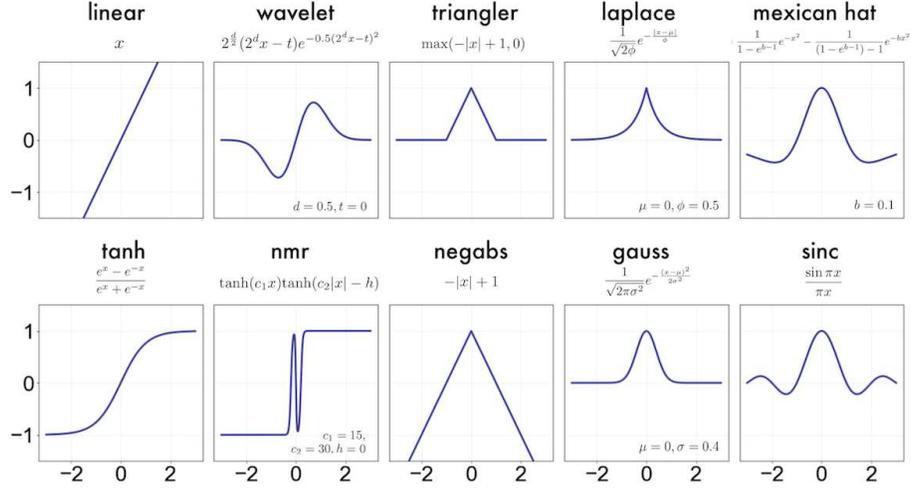}
\caption{10 activation functions applied to ESN.} \label{fig1}
\end{figure}

\vspace{-1mm}
Regarding the activation function $f$ in (\ref{rule}), a monotonic function such as $f(\cdot)=\tanh(\cdot)$ and linear function $f(x)=x$ are conventionally used. 
In addition, we compare wavelet \cite{wang2006harnessing}, non-monotonic response(nmr) \cite{woodward2011reservoir}, and also some even functions. The 
definition of activation functions used in this study are shown in Fig.  \ref{fig1} together with the shape of the graphs. 

\vspace{-2mm}

\section{Experimental settings}\vspace{-2mm}
To compare the prediction accuracy with regards to types of activation functions, 
we consider the following models for generating time series and tasks. \vspace{-3mm}
\paragraph{1. Chaotic time series:}
Two chaotic time series derived from nonlinear functions, i.e. (i) Logistic map: $x(t+1)=4x(t)(1-x(t))$ and (ii) Mackey-Glass equation (MGE): $y(t+1) =0.2y(t-17)/(1+y(t-17))^{10}-0.1y(t)$, 
are considered. 
For a normal prediction task, $y(t)$ is predicted by the trained ESN with the input $y(t-1)$ for Logistic map and $y(t-10)$ for MGE. 
We also test a free running prediction task in the case of MGE, where the input to the trained ESN is $\hat{y}(t-1)$. $\hat{y}$ means predicted $y$. \vspace{-3mm}
\paragraph{2. NARMA time series: }
Nonlinear autoregressive moving average (NARMA) is defined as
$
y(t) = 0.3 y(t-1) + 0.05 y(t-1)\sum_{i=1}^ny(t-i) + 1.5 u(t-n)u(t-1) + 0.1$, 
where $u(t)$ is drawn from a uniform distribution  in the interval $[-1, 1]$. 
We consider the cases of $n=10$ and $n=20$. In the case of $n=20$, $\tanh$ is 
applied to the left hand side of the model. 
For a normal prediction task, the input in (\ref{rule}) is the same as $u(t)$ of the model.

The parameter for ESN is that the spectral radius of $W$ is scaled to $0.95$, and the initial states for all units are 0. The error is evaluated by logarithm of the normalized mean-square error (logNMSE):
$
E(y,\hat{y}) = \log(\langle \|\hat{y} - y  \|^2  \rangle/\langle \|y - \langle y \rangle  \|^2\rangle),  
$
where $\langle \cdot \rangle$ represents time average.

\begin{figure}[t]
\includegraphics[width=\textwidth]{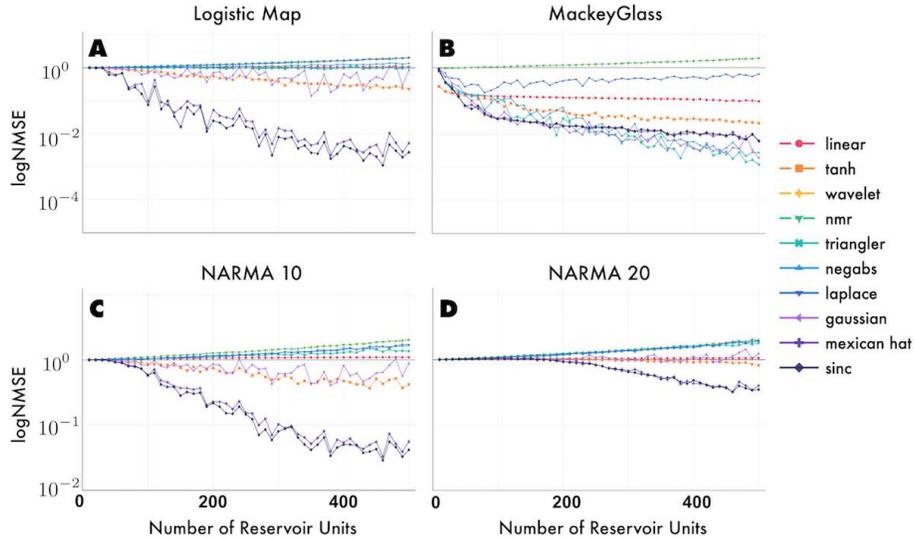}
\caption{Dependency of the prediction accuracy on the size of ESN for a standard prediction task. 
A: Logistic map, B: Mackey-Glass, C: NARMA10, and D: NARMA20 time series.  
$\sigma=0.01$} \label{fig2}
\end{figure}
\vspace{-2mm}
\section{Results and Discussion}\vspace{-2mm}
Figure \ref{fig2} shows the dependency of the prediction accuracy on the size of ESN with respect to ten activation functions in Fig. \ref{fig1} for the normal prediction task. The accuracy for each parameter is the median of 25 trials. It is clearly observed that  in all time series, the sinc and the mexican hat functions perform better than other activation functions at sufficiently large size of ESN. 
Specifically, in the case of Mackey Glass time series, the even functions (except Laplace) predict  better than the odd functions. 

\begin{figure}[t]
\includegraphics[width=\textwidth]{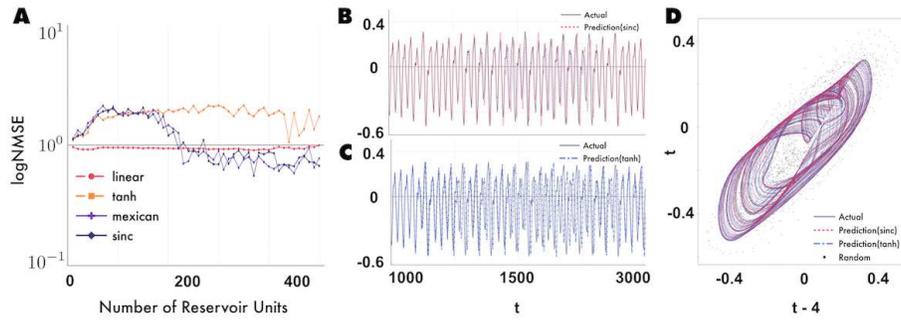}
\caption{A: Dependency of the prediction accuracy on the size of ESN 
 for a free running prediction task. 
B,C: Time series of target and free running predicted signal for the $\text{sinc}$ (B) and $\tanh$ (C) functions. 
D: Reconstructed attractors in delay coordinates from target, free running predicted signal, and random shuffle surrogate time series. $\sigma=0.1$. } \label{fig3}
\end{figure}

Figure \ref{fig3} A shows the dependency of the prediction accuracy on the size of ESN with respect to four activation functions for the free running prediction task.  Similar to the normal prediction, the accuracy of the sinc  function is better than that of the tanh function. However, the values of accuracy is not as high as the normal prediction task in Fig. \ref{fig2}. This is obviously because of the sensitive dependency on initial conditions as shown in Fig. \ref{fig3} B and C. On the other hand, due to the reconstructed attractors from the time series as shown in Fig. \ref{fig3} D, attractors for the predicted time series are closer to the target compared to the random shuffle surrogate. 
In fact, we calculate the distance between the attractors by the Wasserstein metric \cite{muskulus2011wasserstein}, it is possible to confirm that the predicted attractors are quantitatively closer to the target one than the random attractor.  This implies that ESNs with both the sinc and tanh functions learn the target at the level of attractors.

According to the above observations, we discuss the effect of the sinc function for activation as follows.  (1) From the normal prediction, intermediately nonlinear activation function makes the distribution of 
units states more various than the less nonlinear function. Consequently, nonlinearity and  smoothness of the function may increase the predictability. (2) From the free running prediction, more nonlinear function is able to predict longer future, which implies that some specific nonlinear shape of function may increase memory capability.

\vspace{-4mm}

%
%
%
\bibliographystyle{splncs04}

\end{document}